\documentclass[twoside,11pt]{article}

\usepackage{jmlr2e}
\usepackage{amssymb}
\usepackage{amsfonts}
\usepackage{amsmath}
\usepackage{dsfont}
\usepackage{psfrag}
\usepackage{rotating}
\usepackage{url}
\usepackage{bm}

\usepackage{algorithm}
\usepackage{algorithmic}

\usepackage{algorithm}  
\usepackage{algorithmic}  

\usepackage{color,soul}
\usepackage{lscape}
\usepackage{graphicx,epstopdf}

\usepackage{multirow}

\usepackage{graphicx,epstopdf} % more modern

\usepackage{subfigure}

\usepackage{natbib}

\usepackage{algorithm}
\usepackage{algorithmic}

\usepackage{hyperref}

\newcommand{\Proj}{\mathcal{P}}
\newcommand{\loss}{e}
\newcommand{\Loss}{E}
\newcommand{\RRe}{\mathbb{R}}
\newcommand{\trace}[1]{\textrm{tr}\left(#1 \right)}
\newcommand{\Frob}{{2,2}}
\newcommand{\Sparse}{{1,1}}
\ShortHeadings{Overlapping Trace Norms in Multi-View Learning}{Behmardi, Archambeau, and Bouchard}
\firstpageno{1}

\begin{document}

\title{Overlapping Trace Norms in Multi-View Learning}

\author{\name Behrouz Behmardi \email behmaardi@gmail.com \\
       \addr Xerox Research Center Europe\\
       38240 Meylan , France
       \AND
       \name Cedric Archambeau\email cedric.p.archambeau@gmail.com \\
       \addr University College London\\
       Gower Street, London WC1E 6BT, United Kingdom
			 \AND
       \name Guillaume Bouchard \email guillaume.bouchard@xerox.com \\
       \addr Xerox Research Center Europe\\
       38240 Meylan , France}

\editor{--}

\maketitle

\begin{abstract}%   <- trailing '%' for backward compatibility of .sty file
Multi-view learning leverages correlations between different sources of data 
to make predictions in one view based on observations in another view.
A popular approach is to assume that, both, the correlations between the 
views and the view-specific covariances have a low-rank structure,
leading to \emph{inter-battery factor analysis}, a model closely related to \emph{canonical correlation analysis}.
We propose a convex relaxation of this model using
structured norm regularization. Further, we extend the convex formulation to  
a robust version by adding an $\ell_1$-penalized
matrix to our estimator, similarly to convex robust PCA.
We develop and compare scalable algorithms for several convex multi-view
models. We show experimentally that the view-specific correlations are
improving data imputation performances, as well as labeling accuracy in real-world multi-label prediction tasks. 
\end{abstract}


\section{Introduction}

Canonical correlation analysis (CCA) was first introduced by
\cite{Hotelling:Bio36} to analyse the linear relationship between a pair of
random vectors. Similar to principal component analysis (PCA)~\cite{Jolliffe:PCA86}, 
which finds for a random vector a basis in which the
covariance matrix is diagonal and the variances on the diagonal are maximized,
CCA finds two bases in which the correlation matrix between the variables is
diagonal and the correlations on the diagonal are maximized. The principal
directions in PCA span a subspace in which the observations can be projected
with minimum reconstruction error (when defined in terms of sum of squares),
while the canonical directions span a \emph{pair} of subspaces in which the
co-occurring observations are maximally aligned after projection. CCA and its
extensions are widely used in multivariate statistics and machine learning and
have found applications in image
processing~\cite{Borga:LMSP98}, biology~\cite{Parkhomenko:GenBio09} and
geophysics~\cite{Cannon:GeoP08}.

An important limitation of CCA is that it is only able to capture linear
dependencies between two data sets. In order to capture nonlinear alignments,
kernel CCA was introduced~\cite{Bach:JMLR02,Hardoon:Neu04}. In this setting, the
data sets are projected into high-dimensional feature spaces, where a perfect
alignment can always be recovered if the dimension of the feature spaces is sufficiently large. 
Another line of research was triggered by the
probabilistic reformulation of CCA by \cite{Bach:TechRep05}, which in turn led
to robust~\cite{Archambeau:ICML06} and sparse extensions
\cite{Archambeau:NIPS09,Jia:NIPS10}.

Probabilistic CCA assumes that the pair of high-dimensional random vectors are
coupled by a shared low-dimensional latent vector, which captures the
correlations between the two realisations. It provides a natural
framework for generalizing CCA to multiple co-occurring data sets, often called
views, and for accounting for view-specific variations by augmenting CCA with
view-specific low-dimensional latent vectors able to capture the residual
covariance in each view. These two variants have often been loosely called 
CCA by the machine learning community. An example of the former to support
text categorisation in the presence of multi-lingual documents is proposed
by~\cite{Amini:NIPS10}. The latter is known as inter-battery factor analysis
(IBFA)~\cite{Tucker:Psy58,Klami:JMLR13} in the statistics literature.
The main challenge with IBFA is the identifiability of view-specific
and shared variations. This problem was elegantly addressed by
\cite{Virtanen:ICML11,Klami:JMLR13} by proposing a Bayesian treatment of IBFA,
involving group sparsity and on automatic relevance
determination~\cite{MacKay:NeuNet94} to infer
the dimension of the latent vectors (shared and view-specific)
without having to resort too expensive cross-validation.
 
Recently, convex formulations of multi-view learning were also
proposed~\cite{Goldberg:NIPS10,White:NIPS12,christoudias2012multi,cabral2011matrix}, 
extending recent work done on matrix completion, often justified as convex relaxation 
of low-rank recovery problems through the nuclear norm~\cite{Candes:Math09}. To the best of our
knowledge, none of the existing convex multi-view
approaches consider the view-specific variations as part of their model. In some
settings the view-specific covariances could be diagonalized beforehand, but
this normalization step, known as whitening, can be computationally expensive in
high dimension and it is not straightforward in the presence of missing values.
Moreover, a whitening pre-processing step would not take into account possible
linear dependencies between the two data sets and would thus be suboptimal.

We present a convex relaxation of IBFA in the presence of missing values,
which can straightforwardly be generalized to its multi-view version and which
can deal with continuous and discrete data.
The core idea of our approach is to include a
nuclear norm penalty in the objective function for each view to capture the view-specific covariances,
along with a shared nuclear norm penalty to capture the correlations between the
views. We further extend our approach to the robust PCA framework of
\cite{candes2011robust} to account for atypical observations such as outliers.
 
The paper is organized as follows. Section~\ref{sect: convex CCA}
introduces the convex objective for IBFA as a convex
relaxation of the standard probabilistic model. In Section~\ref{sect: admm}, we
describe an efficient algorithm to minimize the objective, using the Alternating
Direction Method of Multipliers (ADMM)l. In Section~\ref{sect:
expe}, we evaluate the model on a wide variety of data sets, and compare our
proposed ADMM algorithm to several off-the-shelf SDP solvers.

\section{Convex formulations of multi-view matrix completion}
\label{sect: convex CCA}
Let $\{(y_{1i},y_{2i}): y_{1i}\in\RRe^{d_1},y_{2i}\in\RRe^{d_2}\}_{i=1}^n$ be co-occurring data pairs.
The number of features in view 1 and 2 are respectively denoted by $d_1$ and $d_2$. 
The main reason for defining two vectors of observations  rather than a single
concatenated vector in the product space $\RRe^{d_1+d_2}$ is that the nature of
the data in each view might be different. For example, in a multi-lingual text application, the views would represent the
features associated with two distinct languages. Another example is image
labeling, where the first view would correspond to the image signature features
and the second view would encode the image labels. In the remainder, we will
restrict our discussion to two views for clarity of the presentation, but the extension to an arbitrary number of views is straightforward. 

\subsection{Overlapping nuclear norms}

We stack the observations $\{y_{1i}\}_{i=1}^n$ and $\{y_{2i}\}_{i=1}^n$ respectively into the matrices
$Y_1:=\{y_{1ij}\}\in\RRe^{d_1\times n}$ and $Y_2:=\{y_{2ij}\}\in\RRe^{d_2\times
n}$. To compare different multi-view approaches, we will consider predictive
tasks, where the goal is to predict missing elements in matrices $Y_1$ and
$Y_2$. The key hypothesis in multi-view learning is that the dependencies
between the views help predicting the missing entries in view 1 given the observed entries in view 1 \emph{and} 2, and vice versa.
Observations are identified by the sets $\Omega_k=\{(i_{kt},j_{kt})\}$ for $k\in\{1,2\}$. 
Each element $(i_{kt},j_{kt})$ represents a pair of (row,column) indices in the $k$-th view.
Predictions are represented by latent matrices $X_1:=\{x_{1ij}\}\in\RRe^{d_1\times n}$ and
$X_2:=\{x_{2ij}\}\in\RRe^{d_2\times n}$. If the value $y_{kij}$ is not observed, our goal is to find a
method that predicts $y_{kij}$ such that the loss $\loss_{k}(x_{kij};y_{kij})$ is minimized on average.
The view-specific losses $\loss_k:\RRe\times\RRe\mapsto\RRe$ are assumed to be
convex in their first argument. Typical examples include
the squared loss $\loss(x,y)=\frac 12(x-y)^2$ for continuous observations 
and the logistic loss $\loss(x,y)=\log(1+e^{-xy})$ for binary observations, $y\in\{-1,+1\}$.
We also define the cumulative training loss associated to view $k$ as 
$\Loss_k(X_k,Y_k) = \sum_{(i,j)\in\Omega_k} \loss_k(x_{kij},y_{kij})$.

In this paper, we study six convex multi-view matrix completion problems called 
I00, I0R, J00, J0R, JL0 or JLR. The sequence of three letters composing their name has the following meaning:
\begin{itemize}
\item The first character (I or J) indicates if the method treats the views independently or jointly;
\item The second character (L or 0) indicates if the method accounts for
view-specific variations as in IBFA. ``L" denotes low-rank as we consider
nuclear norm penalties. %the nuclear norm penalty leads to a convex relaxation of a low-rank matrix completion problem~\cite{Candes:Math09}.
\item The third character (R or 0) indicates if the method is robust.  
Robustness is ensured by including an $\ell_1$-penalized additional view-specific matrix, as in robust PCA~\cite{candes2011robust}.
\end{itemize}

%%%%%%%%%%%%%%%%%%%%%%%%%%%%%%%%%%%%%%%%%%%%%%%%%%%%%%%%%%%%%%%%%%%%%%%%%%%%%%%%%%
\begin{algorithm}[t]
\caption{ADMM for convex multi-view learning}
\begin{algorithmic}[1]
\STATE Initialize $W^0 = \left\{X_0^0, \{X_k^0,S_k^0,Z_k^0\}_{k=1}^2\right\}$
\STATE Initialize $B^0=\{B_1^0,B_2^0\}$, $\mu^0>0$ and $\rho>1$
\FOR{$t = 1$ to $T$}
\STATE $W^t = ADMM_{inner\ loop}(W^{t-1},B^{t-1},\mu^{t-1})$
\FOR{$k = 1$ to 2}
\STATE $B_k^{t} = B_k^{t-1} -\mu^{t-1}(X_k^t+S_k^t+P_kX_0^t-Z_k^t)$
\ENDFOR
\STATE $\mu^{t} = \mu^{t-1}\rho$
\ENDFOR
\end{algorithmic}\label{alg:augLag}
\end{algorithm}
%%%%%%%%%%%%%%%%%%%%%%%%%%%%%%%%%%%%%%%%%%%%%%%%%%%%%%%%%%%%%%%%%%%%%%%%%%%%%%%%%%

\paragraph{Baseline models}
We describe two baseline methods. The first approach, denoted  I00,  
treats the views as being independent, considering a separate nuclear norm penalty for each view:
\begin{eqnarray}
	\min_{X_k}~~
	\lambda_k \|X_k\|_* + \Loss_k(X_k;Y_k)\ ,
	\quad k=1,2
	\enspace,
\label{eq:I00}
\end{eqnarray}
where $\|\cdot\|_*$ denotes the nuclear norm.

The second baseline method, denoted J00, considers a nuclear norm penalty on the concatenated matrix 
$X_0=[X_1;X_2]\in\RRe^{(d_1+d2)\times n}$. The formulation is the most closely related to CCA as will 
be explained shortly. It leads to the following objective:
\begin{eqnarray}
	\min_{X_0}~~
	\lambda_0 \left\|X_0\right\|_* 
	+ 
	\sum_{k=1}^2 \Loss_k(P_kX_0;Y_k)
\label{eq:J00-bis}
\end{eqnarray}
where $P_k$ is a sub-matrix selection operator, so that $P_1X_0$ is the
$d_1\times n$ matrix composed by the first $d_1$ rows of $X_0$ and $P_2X_0$ is
the $d_2\times n$ matrix composed by the last $d_2$ rows of $X_0$. Here, we
consider a single regularization parameter $\lambda_0$, but in some cases, it
might be beneficial to weigh the loss associated to each view differently. For
example, in an image labeling application,  it might be more important to
predict the labels correctly than the features.

The nuclear norm penalty in (\ref{eq:J00-bis}) applies to $X_0$ such that
the matrix to complete is the concatenated matrix $Y_0=[Y_1;Y_2]$. In contrast to I00 as formulated in (\ref{eq:I00}), this enables
information sharing across views, while preserving a view-specific loss to
handle different data types. J00 has been investigated by
\cite{Goldberg:NIPS10}, where the first view was considered as continuous
(containing features), and the second
view was a binary matrix of labels in a multi-label prediction task. The authors
argue that their approach has the advantage of doing multi-task learning,
while handling missing values. J00 is also very similar to the
objective function of the convex multi-view framework of \cite{White:NIPS12}.
%%%%%%%%%%%%%%%%%%%%%%%%%%%%%%%%%%%%%%%%%%%%%%%%%%%%%%%%%%%%%%%%%%%%%%%%%%%%%%%%%%
\begin{algorithm}[t]
\caption{ADMM inner loop }
\begin{algorithmic}[1]
\FOR{$m=1$ to $M$}
\STATE \small{$X_{0}^{m+1} = \mathcal{D}_{\frac{\lambda_0}{\mu^{t-1}}}\biggl([Z_1^{m}+\frac{B_1^{t-1}}{\mu^{t-1}};Z_2^{m}+\frac{B_2^{t-1}}{\mu^{t-1}}]-[X_1^m+S_1^m;X_2^m+S_2^m]\biggl)$}
\FOR{$k = 1$ to 2}
\STATE $X_k^{m+1} = \mathcal{D}_{\frac{\lambda_k}{\mu^{t-1}}}(Z_k^{m}+\frac{B_k^{t-1}}{\mu^{t-1}}-P_kX_0^{m+1}-S_k^m)$
\STATE $S_k^{m+1} = \mathcal{S}_{\frac{\alpha_k}{\mu^{t-1}}}(Z_k^{m}+\frac{B_k^{t-1}}{\mu^{t-1}}-P_kX_0^{m+1}-X_k^{m+1})$
\STATE \small{$W=\left\{X_0^{m+1}, \{X_k^{m+1},S_k^{m+1},Z_k^{m+1}\}_{k=1}^2\right\}\backslash\{Z_k^{m+1}\}$}
\STATE $Z_k^{m+1} =\arg\min_{Z_k} \mathcal{L}(W,Z_k,B^{t-1},\mu^{t-1})$ \label{min-wr-Z}
\ENDFOR
\ENDFOR
\end{algorithmic}\label{alg:admm}
\end{algorithm}
%%%%%%%%%%%%%%%%%%%%%%%%%%%%%%%%%%%%%%%%%%%%%%%%%%%%%%%%%%%%%%%%%%%%%%%%%%%%%%%%%%
 \paragraph{Convex IBFA}\label{sect: convex IBFA}

In contrast to the previous approaches, IBFA accounts for view-specific 
variations~\cite{Tucker:Psy58,Archambeau:ICML06,Virtanen:ICML11,Klami:JMLR13}. 
This can be incorporated into our multi-view matrix completion 
framework by decomposing each view as the sum of a low rank
 \emph{view-specific} matrix $X_k$, as in I00, and a sub-matrix $P_kX_0$ of the 
 \emph{shared} matrix $X_0$ of size $(d_1+d_2)\times n$, as in J00. 
The objective of the resulting method, denoted JL0, is:
\begin{eqnarray}
&\min_{X_0,X_1,X_2}~~
	\lambda_0 \left\|X_0\right\|_* 
	+
	\lambda_1 \|X_1\|_*
	+
	\lambda_2 \|X_2\|_* 
  +\sum_{k=1}^2 \Loss_k(X_k+P_k X_0;Y_k)
	\enspace .	
\label{eq:JL0}
\end{eqnarray} 
It is convex jointly in $X_0$, $X_1$ and $X_2$. As for many nuclear norm
penalized problems, for sufficiently large regularization parameters, the
matrices $X_1$, $X_2$ and $X_0$ are of low-rank at the minimum of the objective. Next, we show that 
JL0 corresponds to a convex relaxation of IBFA by relating it to its probabilistic reformulation.

Consider the probabilistic formulation of CCA~\cite{Bach:TechRep05}:

\begin{align}
&y_{0i}|z_i
	\sim \mathcal{N}\left(\Lambda z_i , \Psi \right) ,
\quad y_{0i}
	= \left(\begin{array}{c}
	y_{1i}\\
	y_{2i}
	\end{array}\right) , \Psi
	= \left(\begin{array}{cc}
	\Psi_1 & 0\\
	0 & \Psi_2
	\end{array}\right) ,
\end{align}

where $z_i\sim\mathcal{N}(0,I)$ is a low-dimensional shared latent variable. 
By introducing additional view-specific latent variables $u_{ki}\sim\mathcal{N}(0,I)$, 
we recover the probabilistic formulation of IBFA~\cite{Klami:JMLR13}:
\begin{align}
&y_{0i}|z_i,u_{0i}
	\sim \mathcal{N}\left(\Lambda z_i + \Gamma u_{0i} , \sigma^2 I \right) ,\nonumber\\
&u_{0i}
	= \left(\begin{array}{c}
	u_{1i}\\
	u_{2i}
	\end{array}\right) ,
\quad \sigma
	= \left(\begin{array}{c}
	\sigma_1\\
	\sigma_2
	\end{array}\right) .
\end{align}
Integrating out the latent variables leads to a Gaussian marginal with covariance matrix given by
\begin{align*}
\Sigma
	&= {\tiny\left(\begin{array}{cc}
	P_1(\Lambda\Lambda^\top+\Gamma\Gamma^\top)P_1^\top + \sigma_1^2 I  & P_1\Lambda\Lambda^\top P_2^\top\\
	P_2\Lambda\Lambda^\top P_1^\top & P_2(\Lambda\Lambda^\top+\Gamma\Gamma^\top)P_2^\top + \sigma_2^2 I
	\end{array}\right) .}
\end{align*}
Hence, the view-specific covariance matrices (blocks along the diagonal) and the 
correlation matrices (off-diagonal blocks) exhibit a low-rank structure.

Instead of integrating out the latent variables, one can consider a Maximum a Posteriori solution. 
As shown in the supplementary material, jointly minimizing the negative log-likelihood 
$-\log \sum_ip(y_{i1},y_{i2},z_{0i},u_{1i},u_{i2})$ with respect to the parameters $\{\Lambda,\Gamma\}$ 
and the latent variables $\{Z,U_0\}$ is equivalent to the following minimization problem:
\begin{align*}
\min_{X_0,X_1,X_2}\|X_0\|_* + \|X_1\|_* + \|X_2\|_* + 
\frac{1}{2}\|\widetilde{Y_1}-P_1X_0- X_1\|_\Frob^2+ \frac{1}{2}\|\widetilde{Y_2}-P_2X_0- X_2\|_\Frob^2,
\end{align*}
where $X_0 \equiv \widetilde\Lambda Z$, $\widetilde\Lambda \equiv \left[\Lambda_1/\sigma_1 ;\Lambda_2/\sigma_2\right]$, 
$X_k \equiv \widetilde\Gamma_k U_k$, $\widetilde{Y}_k \equiv Y_k/\sigma_k$, and $\widetilde{\Gamma}_k \equiv \Gamma_k/\sigma_k$ for $k\in\{1,2\}$.

\paragraph{Robust Convex IBFA} 

Robust PCA reduces the impact of outliers in factor analysis based matrix completion~\cite{candes2011robust},
leading to the I0R and J0R models described in the supplementary material.
Robustness can be incorporated into our convex IBFA formulation by adding a sparse matrix 
$S_k\in\RRe^{d_k\times n}$ to each latent view representation, leading to the prediction of 
$Y_k$ by $P_kX_0 + X_k + S_k$. We will denote the robust convex IBFA model by JLR. Its objective function is defined as follows:
\begin{eqnarray}
&&\hspace{-0.5cm}\min_{X_0,X_1,X_2,S_1,S_2}~~
	\lambda_0 \left\|X_0\right\|_* 
	+
	\lambda_1 \|X_1\|_*
	+ 
	\lambda_2 \|X_2\|_* 
	+ 
	\alpha_1 \| S_1\|_\Sparse
	+ 
	\alpha_2 \| S_2\|_\Sparse
 	\nonumber\\ 
	&&\hspace{-0.5cm} \quad\quad\qquad\qquad 
	+ \sum_{k=1}^2 \Loss_k(X_k+S_k+P_k X_0;Y_k)
	\enspace ,
\label{eq:JLR}
\end{eqnarray}
where $\| \cdot\|_\Sparse$ is the element-wise $\ell_1$ penalty. The level of
sparsity is controlled by view-specific regularization parameters $\alpha_1$ and
$\alpha_2$. Extreme values $y_{kij}$ will tend to be partly explained
by the additional sparse variables $s_{kji}$. Again, the objective is jointly
convex in all its arguments. The main challenge consists in optimizing the large number of 
regularization parameters. We will discuss this into more detail in Section~\ref{sect: reg params}.

\section{Learning algorithm}\label{sect: admm}
The convex objective (\ref{eq:JLR}) can be optimized using off-the-shelf SDP solvers 
such as SDP3 or SeDuMi \cite{sturm:SeDuMi99}. However, these solvers are computationally 
to expensive when dealing with large-scale problems as they use second order information 
\cite{Cai:SIAM10,Toh:JOpt10}. Hence, we propose to use the Alternating Direction Method 
of Multipliers (ADMM)~\cite{Boyd:ML11}, which results in a scalable algorithm. We focus 
only on (\ref{eq:JLR}), as it encompasses all the other objectives.

ADMM is a variation of the Augmented Lagrangian method in which the Lagrangian function 
is augmented by a quadratic penalty term to increase robustness \cite{Bertsekas:Book82}. 
ADMM ensures the augmented objective remains separable if the original objective was separable 
by considering a sequence of optimizations w.r.t. an adequate split of the variables \citep[see][for more details]{,Boyd:ML11}.

We introduce an auxiliary variable $Z_k$ such that it is constrained to be equal to $X_k+S_k+P_kX_0$. 
The augmented Lagrangian $\mathcal{L}(X_0,\{X_k,S_k,Z_k,B_k\}_{k=1}^2,\mu)$ of this problem can be written as:

\begin{eqnarray}\label{eq:augmented_lagrangian_function}
&&\hspace{-0.5cm}
	\lambda_0 \left\|X_0\right\|_* 
	+
	\lambda_1 \|X_1\|_*
	+
	\lambda_2 \|X_2\|_* 
	+
	\alpha_1 \| S_1\|_\Sparse
	+
	\alpha_2 \| S_2\|_\Sparse 
	+ \sum_{k=1}^2 \Loss_k(Z_k;Y_k) 
	\nonumber\\&&\quad%\qquad%\quad 
	-
	 \sum_{k=1}^2  \trace{B_k^\top(X_k +S_k+P_kX_0-Z_k)}
	+
	\frac{\mu}{2}\sum_{k=1}^2\left\|X_k +S_k+P_kX_0-Z_k\right\|_\Frob^2, 
\end{eqnarray}

where $\|\cdot\|_\Frob$ is the element-wise $\ell_2$ norm (or Frobenius norm).
Parameters $B_k$ and $\mu>0$ are respectively the Lagrange multiplier and the
quadratic penalty parameter.

The ADMM algorithm for solving~(\ref{eq:JLR}) is described in
Algorithm~\ref{alg:augLag}. The minimization of
(\ref{eq:augmented_lagrangian_function}) w.r.t. $X_0$, $X_1$, and $X_2$ is a
soft-thresholding operator on their singular values. It is defined as
$\mathcal{D}_\beta(X) = U(\Sigma-\beta I)_+V^T$ for $X= U\Sigma V^T$ and
$\beta\ge 0$. Similarly, the minimization of
(\ref{eq:augmented_lagrangian_function}) w.r.t. $S_1$ and $S_2$ is a
soft-thresholding operator applied element-wise. It is defined as
$\mathcal{S}_\alpha(x) = \textrm{sgn}(x)\max(|x|-\alpha,0)$.
The inner loop is detailed in
Algorithm~\ref{alg:admm}.
In line~\ref{min-wr-Z}, depending on the type of loss, the optimisation of the augmented
Lagrangian w.r.t. $Z_k$ is different. We provide the update formulas for the squared and the logistic loss, but any convex differentiable loss can be used.

\subsection{Squared loss}

The minimization of the augmented Lagrangian w.r.t. $Z_k$ has a closed-form
solution:
{\small
\begin{align*}
Z_k^*
&=\left(\mathds{1_k}-\frac{\Proj_{\Omega_k}(\mathds{1_k})}{\mu}\right)\times\left(X_k+S_k+P_kX_0-
\frac{B_k}{\mu}+\frac{\Proj_{\Omega_k}(Y_k)}{\mu}\right),
\end{align*}
}
where $\mathds{1_k}$ is a matrix of ones and the projection operator $\Proj_{\Omega}:\RRe^{d\times n}\mapsto\RRe^{d\times n}$ selects the entries in $\Omega$ and sets the others entries to 0.
 
\subsection{Logistic loss}

In the case of the logistic loss, the minimization of the augmented Lagrangian w.r.t. $Z_k$ has no analytical solution. However, around a fixed $\bar{Z}_k$, the logistic loss can be upper-bounded by a quadratic function:
\begin{align*}
&\sum_{i,j\in\Omega_k}\log (1+\exp{(-(x_{0ij}+x_{kij}+s_{kij})y_{kij})})\nonumber\\
&\quad\le \frac{\tau}{2}\left\|\Proj_{\Omega_k}(X_k+S_k+P_kX_0)-\Proj_{\Omega_k}(\bar{Y}_k))\right\|_2^2 \enspace,
\end{align*}

where $\bar{y}_{2ij} =
\bar{z}_{kij}-\frac{1}{\tau}\frac{-y_{2ij}}{1+\exp{(y_{2ij}\bar{z}_{kij})}}$ and $\tau$ is the Lipschitz
continuity of the logistic function. This leads to the following solution:
\begin{align*}
Z_k^* =
&\left(\mathds{1}_k-\frac{\Proj_{\Omega_k}(\mathds{1}_k)}{\mu}\right) 
\times\left(X_k+S_k+P_kX_0 - \frac{B_k}{\mu}+\frac{\Proj_{\Omega_k}(\bar{Y}_k)}{\mu}\right) .
\end{align*}
Parameter $1/\tau$ plays the role of a step size~\cite{Toh:JOpt10}. In practice, it can be increased as long as the bound inequality holds. A line search is then 
used to find a smaller value for $\tau$ satisfying the inequality.

%%%%%%%%%%%%%%%%%%%%%%%%%%%%%%%%%%%%%%%%%%%%%%%%%%%%%%%%%%%%%%%%%%%%%%%%%%%%%%%%%%%%%%%%%%%%%%%%%%%%%%%%%%%%%%%%%%
\begin{figure}
\centering
\includegraphics[scale=0.5]{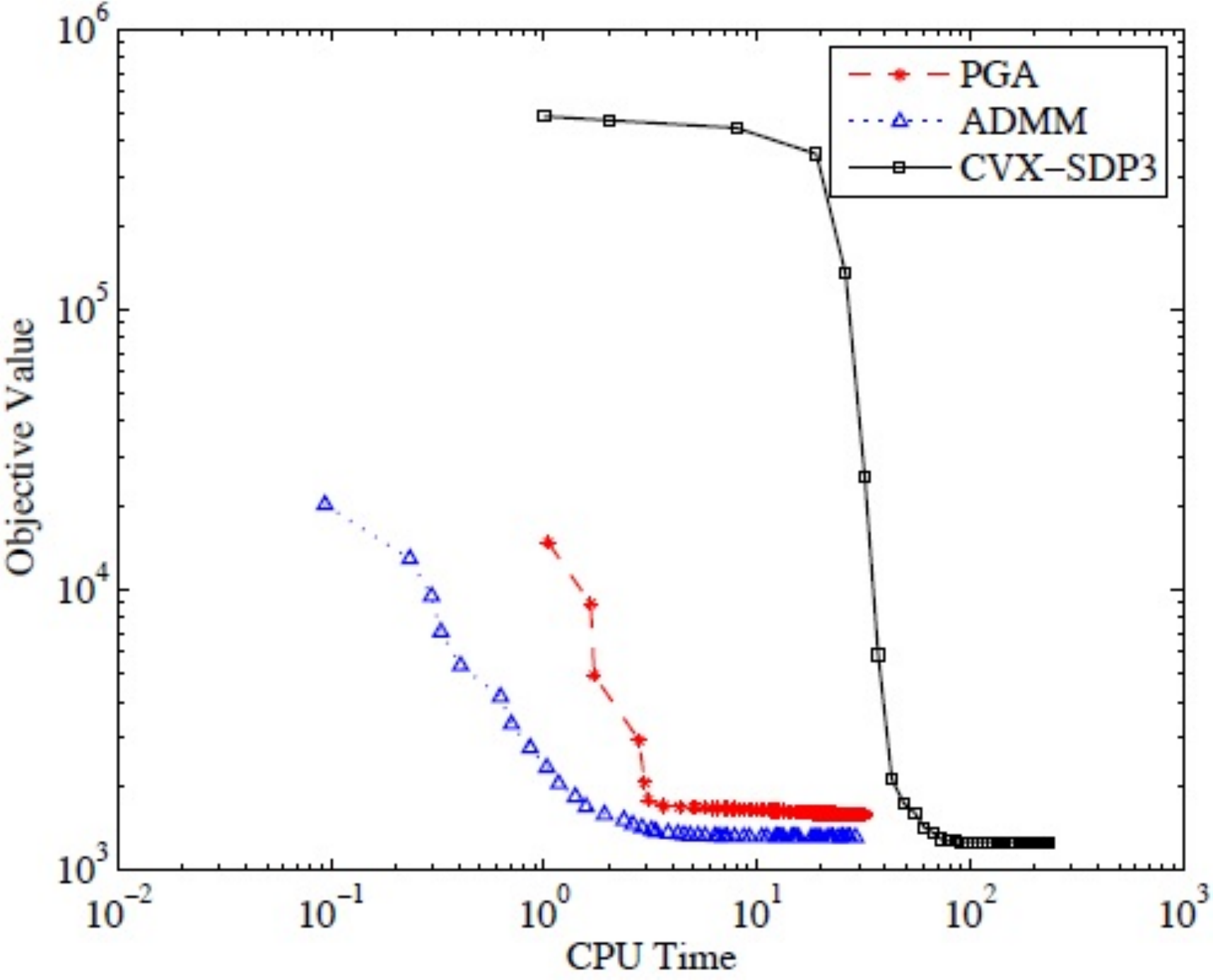}
\caption{Comparison of ADMM, PGA, and CVX with SDP3 solver. ADMM converges faster.}
\label{cvx_adm_pga_comparsion}
\end{figure}
%%%%%%%%%%%%%%%%%%%%%%%%%%%%%%%%%%%%%%%%%%%%%%%%%%%%%%%%%%%%%%%%%%%%%%%%%%%%%%%%%%%%%%%%%%%%%%%%%%%%%%%%%%%%%%%%%%
\subsection{Choice of the regularization parameters}\label{sect: reg params}

The proposed convex minimization algorithm is fast and scalable, but it requires to set a relatively large number of regularization parameters. As with many structured sparsity formulations, a grid search on the held-out dataset performance is not practical and time-consuming. To speed up the adjustment of the regularization parameters, we used Gradient Free Optimization (GFO) , which minimizes the prediction loss on a held-out fraction of the training set. GFO requires only to compute the objective function, but it does not require gradient information. The objective function of GFO is a black-box that takes as input the regularization parameters to optimize. A review of algorithms and comparison of software implementations for a variety of GFO approaches is provided in~\cite{rios2012derivative}.

\section{Experiments}\label{sect: expe}

We evaluate the performance of JLR, JL0, J0R, I0R, and J00 on synthetic data for
matrix completion and on real-world data for image denoising and multi-label
classification. In the following, we first explain the parameter tuning and
evaluation criteria used in the experiments, and then discuss the results.

We used 5-fold cross-validation to obtain the optimum values for the regularization parameters. However, to simplify the optimization, we consider a slightly different formulation of the models. For example for JLR, we optimize w.r.t. $\lambda$ and $c$ where $0<c<1$ instead $\lambda_0$, $\lambda_1$ and $\lambda_2$. The resulting objective function is of the form $\lambda\left(\frac{1}{1-c}\|X_0\|_* + \frac{1}{c}\|X_1\|_* + \frac{1}{c}\|X_2\|_*\right)$. For $\lambda$, $\alpha_1$ and $\alpha_2$, we consider a range of $\{10^{-2},10^{-1},10^0,10^1,10^2\}$ and for $c$ we consider $c = \{0.1,0.2,\ldots,0.9\}$.

To evaluate the performance on the matrix completion problem, we use normalized prediction
test error (which we call test error). We use one part of the data to train the models and test on the remaining part. For multi-label classification experiment, we use the same criteria as
\cite{Goldberg:NIPS10}, that is, the transductive label error. It corresponds to the percentage
of incorrectly predicted labels and the relative feature reconstruction error.

\subsection{Comparison of the solvers}\label{sect:algorithm_comparison}

As a sanity check, we compared our ADMM algorithm for JLR to off-the-shelf SDP solvers (using CVX with SDP3).
We also implemented an alternative method based on the accelerated proximal gradient (PGA)~\cite{Toh:JOpt10}
and we compared to it (see the appendix for a detailed description of PGA).
We report the CPU runtime and the objective. We consider synthetically generated data with $n=100$. We compute the CPU time using the built-in function in MATLAB \textit{cputime}. All algorithms run on a standard desktop computer with $2.5$ GHz CPU (dual core)
and $4$ GB of memory. Figure~\ref{cvx_adm_pga_comparsion} shows the results. It can be observed that ADMM is
converging faster to the optimum value (the one found by CVX with duality gap $
< 10^{-8}$) than CVX and PGA. This illustrates the fact that, under general conditions when $\{\mu_k\}$ is an increasing unbounded sequence, and the objective function and constraints are both differentiable, ADMM converges to the optimum solution super Q-linearly like the Augmented Lagrangian Method~\cite{Bertsekas:Book82}. An additional advantage of ADMM is that the optimal step size is just the penalty term $\mu_k$ which makes the algorithm free of tuning parameters, unlike iterative
thresholding algorithms; PGA and other thresholding algorithms are only sub-linear in
theory \cite{Lin:TechRep09}.

\subsection{Synthetic data experiments}

We compare the prediction capabilities of JLR, JL0, J0R, J00, and I00 on
synthetic datasets. We use randomly generated square matrices of size $n$ for
our experiment. We generate matrices $X_0$, $X_1$, and $X_2$ with different rank
($r_0$, $r_1$, and $r_2$) as a product of $UV^T$ where $U$ and $V$ are generated
randomly with Gaussian distribution and unitary noise. The noise matrices $E_1$
and $E_2$ are generated randomly with Gaussian distribution and unitary noise.
The sparse matrices $S_1$ and $S_2$ are generated by choosing a sparse support
set of size $k=0.1*n^2$ uniformly at random, and whose non-zero entries are
generated uniformly in $[-a,a]$. For each setting, we repeat $10$ trials and
report the mean and the standard deviation of the test error.
%%%%%%%%%%%%%%%%%%%%%%%%%%%%%%%%%%%%%%%%%%%%%%%%%%%%%%%%%%%%%%%%%%%%%%%%%%%%%%%%%%%%%%%%%%%%
\begin{table}%
\centering
%\begin{tabular}{p{.6cm}*{3}{|@{\ }p{1.8cm}@{\ }}}
\begin{tabular}{cccc}
%Meth. 
&  test error & training loss & CPU time\\\hline 
JLR & $83.61\pm 4.19$ & $79.29\pm 2.08$ & $172.24\pm 1.92$\\
JL0 & $89.52\pm 3.97$ & $81.59\pm 2.14$ & $149.14\pm 1.19$\\
J0R & $94.51\pm 4.02$ & $85.37\pm 2.49$ & $89.54\pm 2.01$\\
J00 & $138.31\pm 3.92$ & $92.47\pm 2.84$ & $45.34\pm 1.21$\\
I0R & $131.43\pm 3.16$ & $86.37\pm 2.59$ & $41.26\pm 1.32$\\
I00 & $142.57\pm 3.25$ & $98.61\pm 3.31$ & $46.82\pm 1.52$\\
\end{tabular}
%\caption{
%Test error performance for the synthetic datasets where $n=2000$, $d_1=d_2=1000$.}
\caption{
Performance on the synthetic datasets where $n=2000$, $d_1=d_2=1000$.}
\label{tb:MSE_for_all_approaches_n2000}
\end{table}
%%%%%%%%%%%%%%%%%%%%%%%%%%%%%%%%%%%%%%%%%%%%%%%%%%%%%%%%%%%%%%%%%%%%%%%%%%%%%%%%%%%%%
%%%%%%%%%%%%%%%%%%%%%%%%%%%%%%%%%%%%%%%%%%%%%%%%%%%%%%%%%%%%%%%%%%%%%%%%%%%%%%%%%%

\begin{table}
\centering
%\begin{tabular}{p{.6cm}*{3}{|@{\ }p{1.8cm}@{\ }}}
\begin{tabular}{cccc}
%Meth.
&  test error &  training loss & CPU time\\\hline 
JLR &$53.23\pm 2.25$ & $27.43\pm 0.12$ & $3.96\pm 0.19$\\
JL0 &  $59.45\pm 2.74$ & $28.73\pm 0.15$ & $2.71\pm 0.14$\\
J0R & $63.56\pm 2.17$ & $29.76\pm 0.22$ & $1.08\pm 0.21$ \\
J00 &  $72.39\pm 2.32$ & $45.37\pm 0.31$ & $1.03\pm 0.29$ \\
I0R &  $69.25\pm 2.13$ & $41.52\pm 0.23$ & $1.02\pm 0.36$ \\
I00 &  $76.41\pm 2.57$ & $48.36\pm 0.27$ & $1.01\pm 0.31$ \\
\end{tabular}
%\caption{Test error performance for the synthetic datasets where $n=200$, $d_1=d_2=100$.}
\caption{Performance on the synthetic datasets where $n=200$, $d_1=d_2=100$.}
\label{tb:MSE_for_all_approaches_n200}
\end{table}

%%%%%%%%%%%%%%%%%%%%%%%%%%%%%%%%%%%%%%%%%%%%%%%%%%%%%%%%%%%%%%%%%%%%%%%%%%%%%%%%%%%%%%%%%%%%%%%%%%%%%%%%%%%%%%%%%%

Tables~\ref{tb:MSE_for_all_approaches_n2000} and~\ref{tb:MSE_for_all_approaches_n200} 
show the comparison of the approaches discussed in Section~\ref{sect: convex CCA} for two different settings. 
Each cell shows the mean and standard deviation of the test error over $10$ simulations. 
The test prediction performances of JLR is superior compared to the other approaches. 
We also see that the training loss is lower in JLR approach. Note that the stopping criteria is a fixed number
of iterations, and more complex take more time than simpler ones, but the loss of speed is only by a constant amout
(JLR is about 3 times slower to learn than I00).

\subsection{Image denoising}

We evaluate the performance of JLR for image denoising and compare it against
J0R and I0R. The image denoising is based on the Extended Yale Face Database
B available at {\small \url{cvc.yale.edu/projects/yalefacesB.html}}. It
contains image faces from 28 individuals under 9 different poses and 64
different lighting conditions. We choose two different lighting conditions
(+000E+00 and +000E+20) as two views of an image. The intuition is that each
view has low rank latent structure (due to the view-specific lightning condition),
while each image share the same global structure (the same person with the same pose).
Each image is down-sampled to $100\times 100$. So the
dimension of the datasets based on our notation is:  $d_1=10000$, and
$d_2=10000$. We add $5\%$ noise to the randomly selected pixels of view 1 and
view 2 as well as 40\% of missing entries in both views. The goal is to reconstruct the
image by filling in missing entries as well as removing the noise.

%%%%%%%%%%%%%%%%%%%%%%%%%%%%%%%%%%%%%%%%%%%%%%%%%%%%%%%%%%%%%%%%%%%%%%%%%%%%%%%%%%%%%%%%%%%%%%%%%%%%%%%%%%%%%%%%%%

\begin{figure*}[t]
\centering
\subfigure[Original]{
\resizebox{2cm}{!}{
\includegraphics{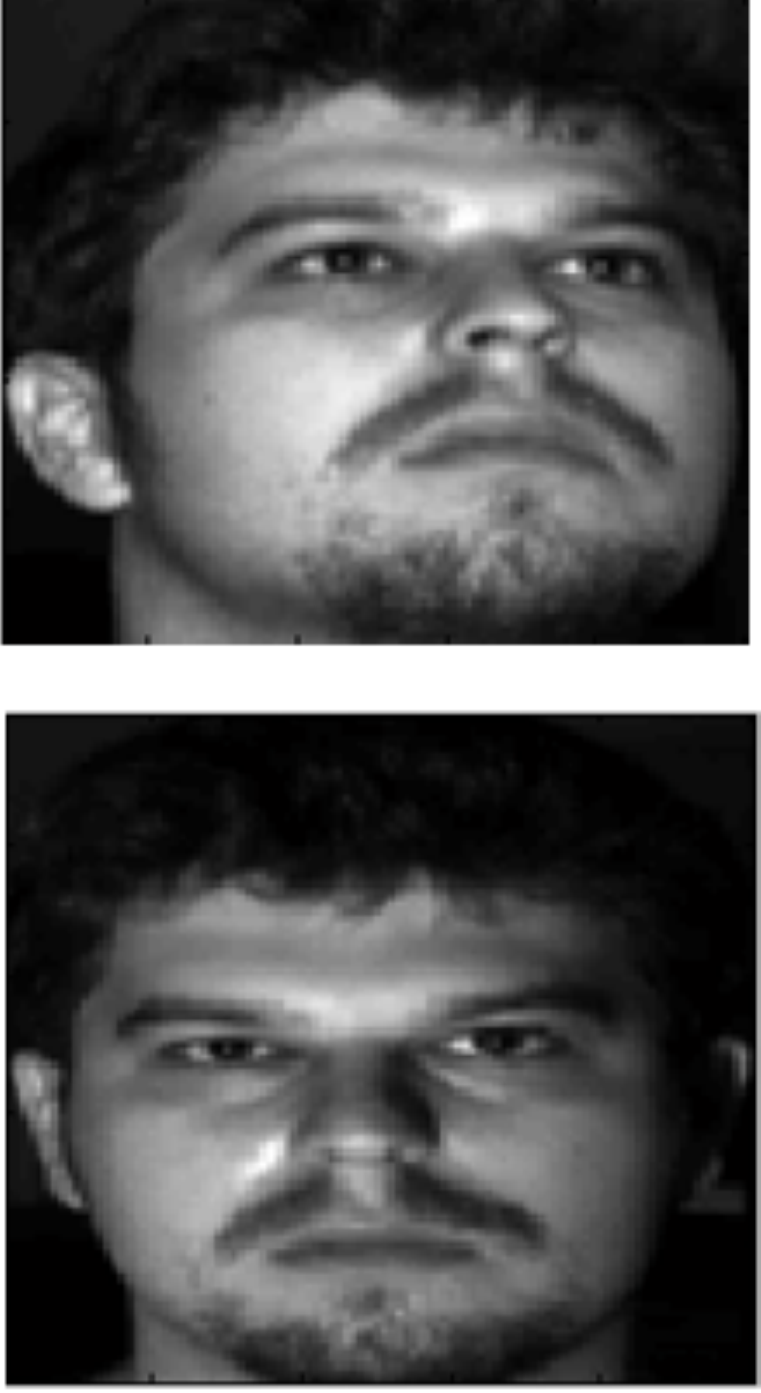}}
}
\subfigure[Corrupted]{
\resizebox{2cm}{!}{
\includegraphics{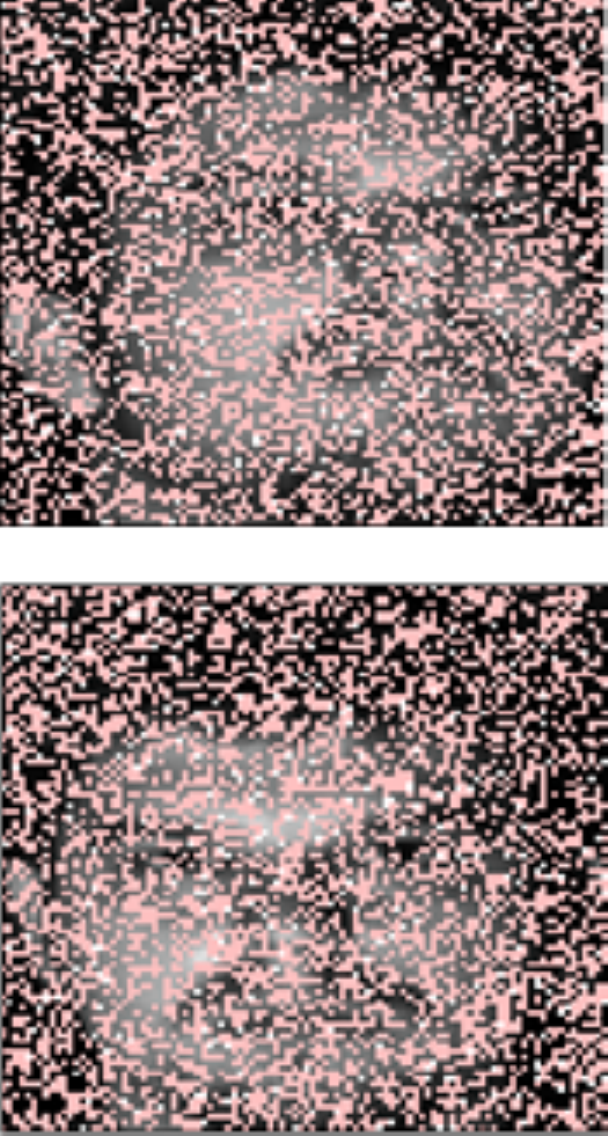}}
}
\subfigure[JLR]{
\resizebox{2cm}{!}{
\includegraphics{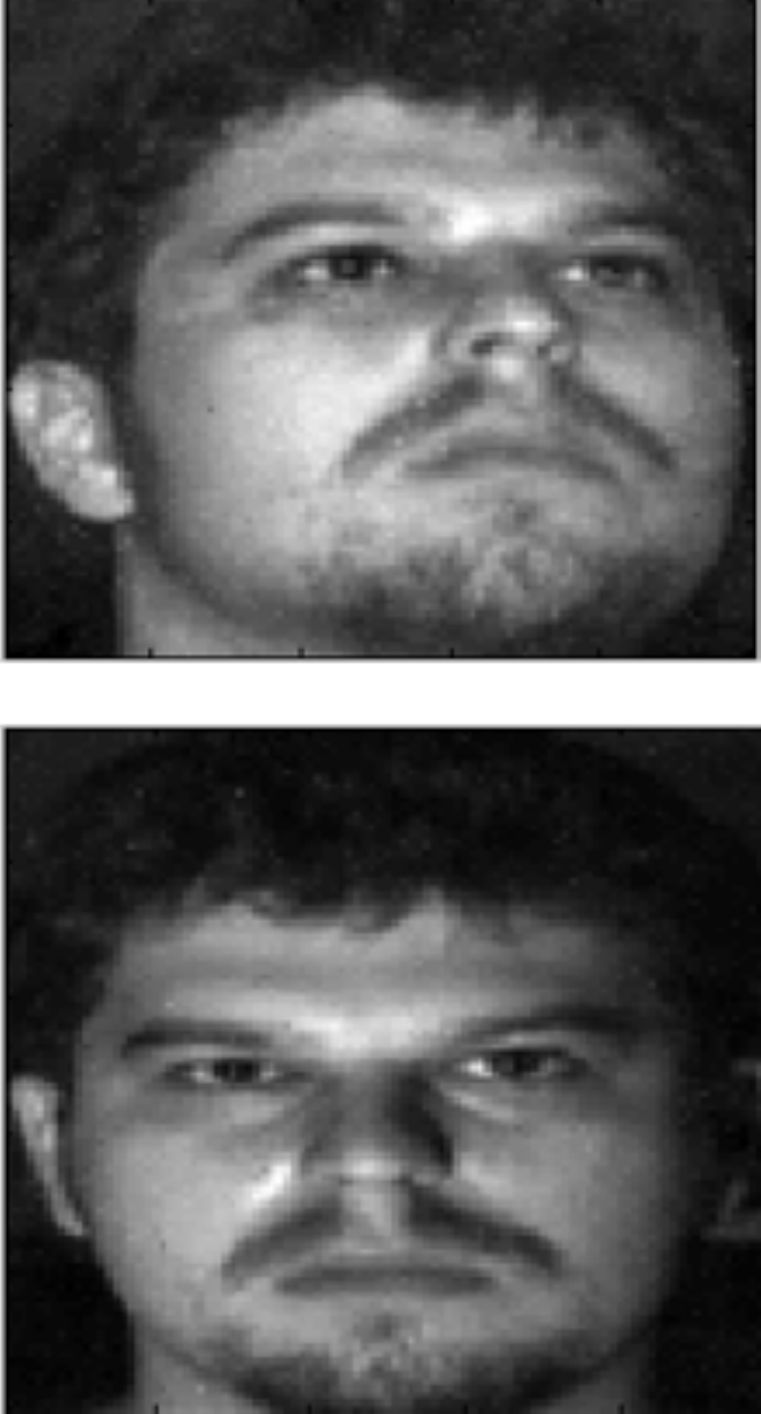}}
}
\subfigure[J0R]{
\resizebox{2cm}{!}{
\includegraphics{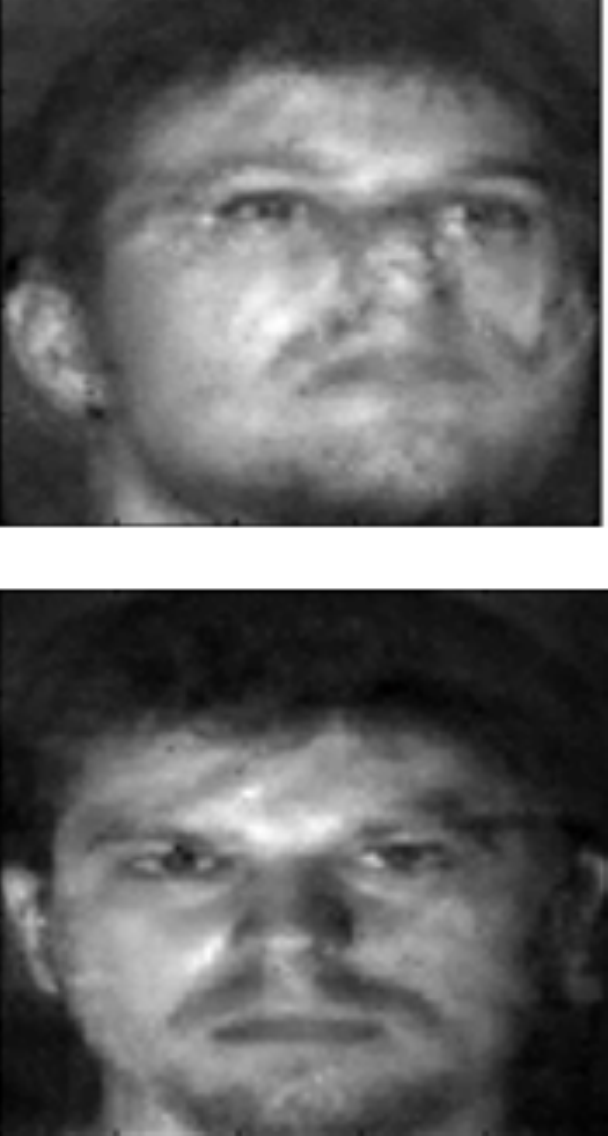}}
}
\subfigure[I0R]{
\resizebox{2cm}{!}{
\includegraphics{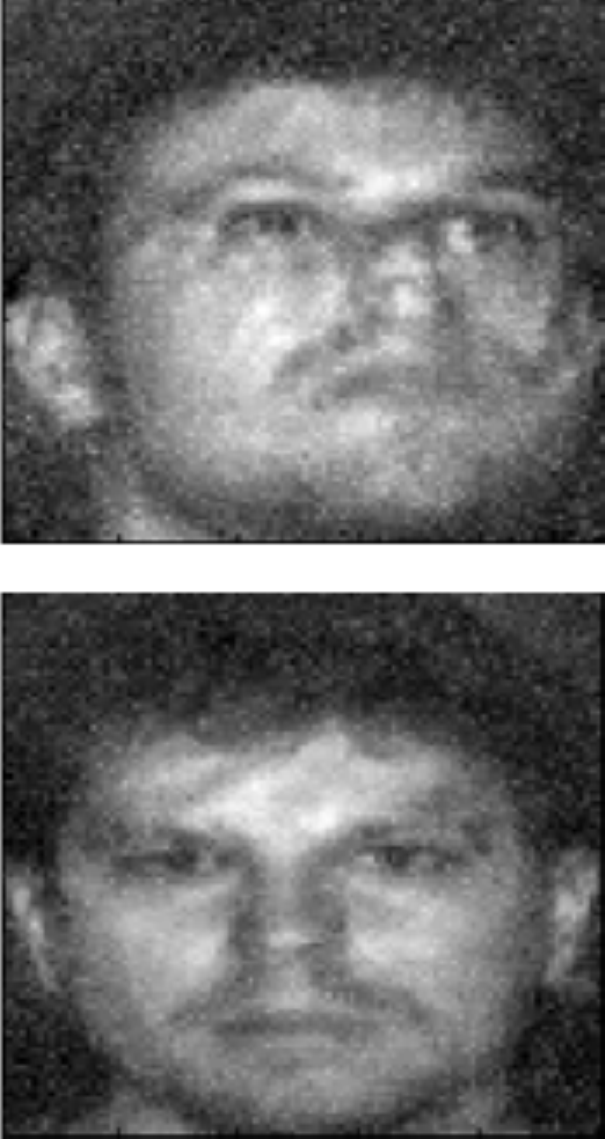}}
}
\caption{Reconstruction of a noisy image with $5\%$ uniform noise and $40\%$ missing entries at random. The reconstruction of noisy images by JLR is qualitatively better than the reconstruction by the other approaches. This is confirmed quantitatively in Table~\ref{tb:MSE_for_face_image}.}\label{im:image_reconstruction_5per_noise}
\end{figure*}
%%%%%%%%%%%%%%%%%%%%%%%%%%%%%%%%%%%%%%%%%%%%%%%%%%%%%%%%%%%%%%%%%%%%%%%%%%%%%%%%%%%%%%%%%%%%%%%%%%%%%%%%%%%%%%%%%%

%%%%%%%%%%%%%%%%%%%%%%%%%%%%%%%%%%%%%%%%%%%%%%%%%%%%%%%%%%%%%%%%%%%%%%%%%%%%%%%%%%%%%%%%%%%%%%%%%%%%%%%%%%%%%%%%%%
{\small
\begin{table}[t]
\centering
\small
\begin{tabular}{cccccc}%{p{0.8cm}p{0.8cm}p{0.8cm}p{0.8cm}p{0.8cm}p{0.8cm}}

   JLR & JL0 & J0R & J00 & I0R & I00 \\\hline\\ [-1.5ex]
  $0.0619$ & $0.0621$ & $0.0631$& $0.1002$ & $0.0825$ & $0.1012$ \\%\hline
\end{tabular}
\caption{
Average test error (squared error) over $5$ random train-test splits for the Yale Face Dataset. 
The standard deviation was less than $10^{-3}$ in all cases.}\label{tb:MSE_for_face_image}
\end{table} 
}
%%%%%%%%%%%%%%%%%%%%%%%%%%%%%%%%%%%%%%%%%%%%%%%%%%%%%%%%%%%%%%%%%%%%%%%%%%%%%%%%%%%%%%%%%%%%%%%%%%%%%%%%%%%%%%%%%%

Figure~\ref{im:image_reconstruction_5per_noise} shows the visualization
performance in reconstructing the noisy image. J0R is successful
in removing the noise but the quality of the reconstruction is visually worse than the
JLR models that captures well the specific low-rank variations of each image. We also 
compare the test errors (squared loss) of the different models in Table~\ref{tb:MSE_for_face_image}. 
The visual intuition is confirmed by the fact that the best performances are obtained by JLR and JL0. 
Quantitatively, JLR only slightly outperforms JL0, but there is an important visual qualitative improvement in Figure~\ref{im:image_reconstruction_5per_noise}.

\subsection{Multi-label classification}

We evaluate the applicability of JLR with a logistic loss on the second view in the context of a multi-label prediction task and compare it with the approach proposed by
\cite{Goldberg:NIPS10}. In this application, View~1 represents the feature matrix and View~2 the label
matrix. In many real situations, the feature matrix is partially
observed. One simple solution is to first impute the missing data in the feature 
matrix and then further proceed with the multi-label classification task.
Another way is to treat the feature matrix and the label matrix as two views of
the same object, and treating the labels to predict as missing entries. We also compared these algorithms against the state-of-the-art 
approaches for multi-label classification. We use Robust PCA (using our ADMM algorithm) for data imputation in the feature matrix 
and then use two different approaches in multi-label classification proposed in \cite{zhang2007ml} and linear SVM. 
%
%Note that in \cite{Goldberg:NIPS10}, the proposed approach was compared with several state-of-the-art approach in matrix completion and classification which due to space limitation we just consider the comparison of our approach with the model in \cite{Goldberg:NIPS10}.
We consider two different datasets, both of which were used by
\cite{Goldberg:NIPS10}, namely: Yeast Micro-array data and Music
Emotion data available at the following url: {\small \url{mulan.sourceforge.net/datasets.html}}.

%%%%%%%%%%%%%%%%%%%%%%%%%%%%%%%%%%%%%%%%%%%%%%%%%%%%%%%%%%%%%%%%%%%%%%%%%%%%%%%%%%%%%%%%%%%%%%%%%%%%%%%%%%%%%%%%%%
{\small
\begin{table*}[t]
\centering

\begin{tabular}{lccc}
&\multicolumn{3}{c}{Label error percentage}\\%\hline
& $\pi=40\%$ & $\pi=60\%$ & $\pi=80\%$ \\\hline\\ [-1.5ex]
J00 \cite{Goldberg:NIPS10}&$16.7(0.3)$ & $13.0(0.2)$&$8.5(0.4)$\\
J00 (ADMM) &$16.8(0.4)$ & $13.1(0.2)$&$8.4(0.3)$\\
J0R  &$16.4(0.2)$ & $12.9(0.1)$&$8.1(0.2)$\\
JL0  &$16.8(0.2)$ & $13.0(0.1)$&$8.4(0.3)$\\
JLR  &\textbf{16.4(0.2)} & $\textbf{12.8(0.1)}$&$\textbf{8.1(0.2)}$\\
Imputation + \cite{zhang2007ml} & $20.3(0.2)$ & $19.5(0.1)$&$18.4(0.1)$\\
Imputation + SVM & $21.6(0.3)$ & $20.5(0.1)$&$20.4(0.2)$\\\hline\\
&\multicolumn{3}{c}{Relative feature reconstruction error}\\
& $\pi=40\%$ & $\pi=60\%$ & $\pi=80\%$\\\hline\\
J00 \cite{Goldberg:NIPS10}&$0.86(0.02)$ & $0.92(0.00)$&$0.74(0.02)$\\
J00 (ADMM) &$0.83(0.01)$ & $0.89(0.01)$&$0.71(0.01)$\\
J0R  &$0.81(0.01)$ & $0.86(0.01)$&$0.69(0.01)$\\
JL0  &$0.82(0.01)$ & $0.85(0.01)$&$0.70(0.01)$\\
JLR  &$\textbf{0.74(0.01)}$ & $\textbf{0.82(0.01)}$&$\textbf{0.67(0.01)}$\\
Imputation + \cite{zhang2007ml} &$0.80(0.00)$ & $0.75(0.02)$&$0.74(0.01)$\\
Imputation + SVM &$0.80(0.00)$ & $0.75(0.02)$&$0.74(0.01)$\\\hline\\

\end{tabular}
\caption{Label prediction error comparison for Yeast data, where the first J00 is proposed in~\cite{Goldberg:NIPS10}.}
\label{tb:error_comparison_yeast}
\end{table*}
}
%%%%%%%%%%%%%%%%%%%%%%%%%%%%%%%%%%%%%%%%%%%%%%%%%%%%%%%%%%%%%%%%%%%%%%%%%%%%%%%%%%%%%%%%%%%%%%%%%%%%%%%%%%%%%%%%%%

%%%%%%%%%%%%%%%%%%%%%%%%%%%%%%%%%%%%%%%%%%%%%%%%%%%%%%%%%%%%%%%%%%%%%%%%%%%%%%%%%%%%%%%%%%%%%%%%%%%%%%%%%%%%%%%%%%
{\small
\begin{table*}
\centering
\begin{tabular}{lccc}
&\multicolumn{3}{c}{Label error percentage}\\%\hline
& $\pi=40\%$ & $\pi=60\%$ & $\pi=80\%$\\\hline\\ [-1.5ex]
J00 \cite{Goldberg:NIPS10}&$27.4(0.8)$ & $23.7(1.6)$&$19.8(2.4)$\\
J00 (ADMM) &$28.0(0.01)$ & $24.1(0.02)$&$21.2(0.01)$\\
J0R  &$28.0(0.01)$ & $24.1(0.02)$&$21.2(0.01)$\\
JL0  &$27.8(0.02)$ & $23.0(0.05)$&$20.7(0.06)$\\
JLR  &$27.8(0.02)$ & $\textbf{23.0(0.05)}$&$20.7(0.06)$\\
Imputation + \cite{zhang2007ml} & $\textbf{25.6(1.0)}$ & $23.5(0.9)$&$\textbf{19.5(1.1)}$\\
Imputation + SVM & $26.7(0.7)$ & $25.7(1.1)$&$24.3(1.5)$\\\hline\\

&\multicolumn{3}{c}{Relative feature reconstruction error}\\%\hline
& $\pi=40\%$ & $\pi=60\%$ & $\pi=80\%$\\\hline\\ [-1.5ex]
J00 \cite{Goldberg:NIPS10}&$0.60(0.05)$ & $0.46(0.12)$&$0.25(0.03)$\\
J00 (ADMM) &$0.58(0.01)$ & $0.33(0.02)$&$0.12(0.01)$\\
J0R  &$0.58(0.01)$ & $0.33(0.02)$&$0.12(0.01)$\\
JL0  &$0.56(0.02)$ & $0.30(0.01)$&$0.10(0.01)$\\
JLR  &$0.56(0.02)$ & $\textbf{0.30(0.01)}$&$\textbf{0.10(0.01)}$\\
Imputation + \cite{zhang2007ml} &$\textbf{0.47(0.01)}$ & $0.31(0.02)$&$0.12(0.01)$\\
Imputation + SVM &$0.47(0.01)$ & $0.31(0.02)$&$0.12(0.01)$\\\hline\\
\end{tabular}
\caption{
Label prediction error comparison for music data, where 
the first J00 is copied from~\cite{Goldberg:NIPS10}.}
\label{tb:error_comparison_music_emotion}
\end{table*}
}
%%%%%%%%%%%%%%%%%%%%%%%%%%%%%%%%%%%%%%%%%%%%%%%%%%%%%%%%%%%%%%%%%%%%%%%%%%%%%%%%%%%%%%%%%%%%%%%%%%%%%%%%%%%%%%%%%%

The Yeast dataset contains $n=2417$ samples in $d_1 = 103$
dimensional space. Each sample can belong to one of $d_2 = 14$ gene functional
classes and the goal is to classify each gene based on its function. We vary the
percentage of observed value ($\pi=40\%$, $\pi=60\%$, and $\pi=80\%$). For each
$\pi$, we run $10$ repetitions and report mean and standard deviation (in
parenthesis). Parameters are tuned by cross validation on optimizing the label
error prediction. 
The left columns of Table~\ref{tb:error_comparison_yeast} show the
label prediction error on Yeast dataset. We observe that J00~\cite{Goldberg:NIPS10}
and J00 (ADMM) produce very similar results. We obtain a slightly lower label prediction error for J0R and JLR. 
The right columns in Table~\ref{tb:error_comparison_yeast} show 
the relative feature reconstruction error. 
JLR outperforms other algorithms in relative
feature reconstruction error. This is due to the fact that JLR is a richer model, better able to capture the underlying structure of the data. We compared these algorithms against a baseline in which $100\%$ of features are given (i.e., no missing entries) and predict the missing labels using SVM. The prediction performance for the baseline with $\pi=40\%, 60\%, 80\%$ is $20.9(0.1), 19.4(0.3), 18.8(0.2)$, respectively. A paired $t$-test shows that JLR outperforms the baseline at a significance level of $5\%$.

The Music dataset consists of $n=593$ songs in $d_1=72$ dimension (i.e., $8$
rhythmic and $64$ timbre-based) each one labeled with one or more of $d_2=6$
emotions (i.e., amazed-surprised, happy-pleased, relaxing-calm, quiet-still,
sad-lonely, and angry-fearful). Features are automatically extracted from a
30-second sound clip.
The left columns of Table~\ref{tb:error_comparison_music_emotion}
show the label prediction error. Similar to Yeast, we observe that J00 \cite{Goldberg:NIPS10} and J00 (ADMM) have the same label error
performance. In this dataset, we see that JLR and JL0 produce the same results
which suggests that the low-rank structure defined on the label matrix is sufficient to
improve the prediction performance. The last columns of
Table~\ref{tb:error_comparison_music_emotion} show the relative feature
reconstruction error. First, it should be noted that J00 (ADMM) has better
results in relative feature reconstruction error compared to J00 \cite{Goldberg:NIPS10}. This
suggest the efficiency of ADMM implemented for J00. Moreover, we
observe that JLR outperforms other algorithms in terms of relative
feature reconstruction error. Again, we compared these algorithms against a baseline in which $100\%$ of features are given (i.e., no missing entries) and predict the missing labels using SVM. The prediction performance for the baseline with $\pi=40\%, 60\%, 80\%$ is $25.8(0.4), 24.6(1.2), 24.5(1.3)$, respectively. Using a paired $t$-test, JLR is statistically outperforming the baseline at the level of $5\%$ significance with $\pi=40\%,60\%$.

\subsection{Optimization of regularization parameters}

We checked wether grid search described at the beginning of this section is the best solution to optimize the cross-validation error 
with respect to the choice of the regularization parameters. After running experiments on synthetic data, we found that \textit{S}table \textit{N}oisy \textit{O}ptimization using \textit{B}ranch and \textit{F}it (SNOBFIT) \cite{huyer2008snobfit} is one of the best algorithms among others for optimizing a black-box function both in terms of accuracy and time efficiency (i.e., compare to multilevel coordinate search (MCS) \cite{huyer1999global} and simplex derivative pattern search method (SID-PSM) \cite{custodio2008sid}). SNOBFIT is a Matlab package designed for selecting continuous parameter setting for simulations or experiments. It is based on global and local search by branching and local fit. We used freely available software for SNOBFIT (available at the following url: {\small \url{mat.univie.ac.at/~neum/software/snobfit/}}). We run experiment for Music emotion dataset and compare the results obtained by both SNOBFIT and grid search. Table~\ref{tb:error_comparison_gfo_cv} shows the predictive performances of the method after tuning the regularization parameters for JLR algorithm. The results show better predictive performances for both label prediction error and reconstruction error using SNOBFIT, but the speed was comparable between the two approaches.
%%%%%%%%%%%%%%%%%%%%%%%%%%%%%%%%%%%%%%%%%%%%%%%%%%%%%%%%%%%%%%%%%%%%%%%%%%%%%%%%%%%%%%%%%%%%%%%%%%%%%%%%%%%%%%%%%%
\begin{table*}
\centering
\begin{tabular}{lcccccc}
&\multicolumn{3}{c}{Label error percentage}&\multicolumn{3}{c}{Relative feature reconstruction error}\\%\hline
& $\pi=40\%$ & $\pi=60\%$ & $\pi=80\%$& $\pi=40\%$ & $\pi=60\%$ & $\pi=80\%$\\\hline\\ [-1.5ex]
JLR-SNOBFIT  &$25.9(0.04)$ & $22.1(0.02)$&$19.3(0.03)$&$0.53(0.04)$ & $0.29(0.03)$&$0.10(0.03)$\\  
JLR-GRID  &$27.8(0.02)$ & $23.0(0.05)$&$20.7(0.06)$&$0.56(0.02)$ & $0.30(0.01)$&$0.10(0.01)$\\

%Robust CCA  & & &\\\hline
\end{tabular}
\caption{
Label prediction error comparison of SNOBFIT vs. grid search on the optimization of cross-validation error for music data.}
\label{tb:error_comparison_gfo_cv}
\end{table*}
%%%%%%%%%%%%%%%%%%%%%%%%%%%%%%%%%%%%%%%%%%%%%%%%%%%%%%%%%%%%%%%%%%%%%%%%%%%%%%%%%%%%%%%%%%%%%%%%%%%%%%%%%%%%%%%%%%

\section{Conclusion}

We introduced a convex reformulation of inter-battery factor analysis, a method closely related to canonical correlation analysis. The proposed formulation
enables us to easily handle missing data, to monitor the convergence of the estimation algorithm, and to  develop scalable algorithms by exploiting tools from convex optimization. A natural extension of our work would be to develop stochastic algorithms  to scale to even bigger datasets. On the modeling side, our approach would benefit from theoretical rank-recovery results to identify view-specific and shared latent subspaces. Experimentally, we showed that accounting for view-specific variations can significantly boost performance and that robust approaches are beneficial in practice. For tuning the regularizaiton parameter in our problem, we use gradient free optimization and show the accuracy can be improved. Future work include the use of overlapping trace norms for a broader class of problems, including collective matrix factorization and tensor factorization. 
We are also investigating smarter ways of selecting regularization parameters using generalization bounds or Bayesian approaches.

%%\iffalse
%%The quest for convexity tends to hide another fundamental problem
%%which is the ability to efficiently tune the regularization parameters
%%by reducing the amount of cross-validation. This is a known 
%%open problem with no obvious solution, but it becomes critical in our case, 
%%since one of the proposed models requires at least five parameters to tune. 
%%An interesting direction would be to generalize the recent results for 
%%Bayesian matrix factorization of~\cite{nakajima2011theoretical} to the CCA setting, 
%%enabling automatic parameter tuning and the ability to obtain uncertainty estimates for 
%%the latent factors and the predictions.
%%\else
%A direction for future research would be to deviceefficiently tune the regularization parameters by reducing the amount of cross-validation.
%%\fi

\bibliography{jmlr}

\end{document}